# Differential Evolution Integrated Hybrid Deep Learning Model for Object Detection in Pre-made Dishes

Lujia Lv
College of Computer and Information Science
Southwest University
Chongqing, China
lvlj0109@163.com

Di Wu
College of Computer and Information Science
Southwest University
Chongqing, China
wudi1986@swu.edu.cn

Yangyi Xia
College of Food Science
Southwest University
Chongqing, China
2658355128@qq.com

Jia Wu
College of Food Science
Southwest University
Chongqing, China
2813324899@qq.com

Xiaojing Liu
Chongqing True Flavor Food Co LTD
Chongqing, China
845004549@qq.com

Yi He
William & Mary
Williamsburg, VA, United States of America
yihe@wm.edu

*Abstract*—With the continuous improvement of people's living standards and fast-paced working conditions, pre-made dishes are becoming increasingly popular among families and restaurants due to their advantages of time-saving, convenience, variety, cost-effectiveness, standard quality, etc. Object detection is a key technology for selecting ingredients and evaluating the quality of dishes in the pre-made dishes industry. To date, many object detection approaches have been proposed. However, accurate object detection of pre-made dishes is extremely difficult because of overlapping occlusion of ingredients, similarity of ingredients, and insufficient light in the processing environment. As a result, the recognition scene is relatively complex and thus leads to poor object detection by a single model. To address this issue, this paper proposes a Differential Evolution Integrated Hybrid Deep Learning (DEIHDL) model. The main idea of DEIHDL is three-fold: 1) three YOLO-based and transformer-based base models are developed respectively to increase diversity for detecting objects of pre-made dishes, 2) the three base models are integrated by differential evolution optimized self-adjusting weights, and 3) weighted boxes fusion strategy is employed to score the confidence of the three base models during the integration. As such, DEIHDL possesses the multi-performance originating from the three base models to achieve accurate object detection in complex pre-made dish scenes. Extensive experiments on real datasets demonstrate that the proposed DEIHDL model significantly outperforms the base models in detecting objects of pre-made dishes.

## I. INTRODUCTION

As living standards rise and work environments become more hectic, pre-made dishes are gaining popularity with both households and restaurants, thanks to benefits such as saving time, ease of use, a wide range of options, affordability, and consistent quality.

Artificial Intelligence has been applied to a wide variety of tasks, such as vision tasks [3-7], big data analytics tasks [43-50,52-59], and so on. Object detection is a key technology in the pre-made dishes industry for selecting ingredients and evaluating dish quality. It can evaluate the quality of the dish based on the detection results. In short, it can increase productivity and drive digital transformation in the pre-made dishes industry.

To date, many object detection approaches have been proposed. They can be roughly divided into four categories: large models, two-stage models, single-stage models, and Transformer-based models. Large models can learn complex patterns, but require a significant amount of computational resources. The two-stage model first generates regions, and then adjusts the boundaries through a classification network extracted from the Convolutional Neural Network(CNN) [29]. The single-stage models directly extract features within the network to determine the category and location of objects. The Transformer-based models can simultaneously analyze the relationships between all elements within feature sequences. It helps the model to better understand the global context information. Based on the previous analysis, we conclude that YOLO-based and Transformer-based models are suitable for the pre-made dishes industry.

However, factors such as overlapping ingredients, similarity, and insufficient lighting in the processing environment make object detection in pre-made dishes extremely challenging. Therefore, scene recognition is more complex, resulting in poor detection effect of a single model. We combined the advantages of these models to construct a hybrid integrated model to improve detection accuracy and generalization ability.

Ensemble learning enhances the generalization performance of a single model by integrating the prediction results of multiple learning models. Based on the idea of integration, this paper proposes a Differential Evolutionary Integrated Hybrid Deep Learning (DEIHDL) model. The core concept of DEIHDL includes three points: 1) creating three YOLO-based and Transformer-based basic models to enrich pre-made dishes detection, 2) integrating these models using differential evolution algorithm, and 3) using weighted boxes fusion strategy to evaluate confidence during the integration. This enables DEIHDL to integrate the advantages of multiple models and accurately detect targets in complex pre-made dish scenes.

The main contributions of this paper include:

- A DEIHDL model. It is able to integrate three base models of different types by differential evolution optimized self-adjusting weights.
- The algorithm design and analysis is given for the proposed DEIHDL model.
- Extensive experiments were conducted on the real-world dataset and the results were analyzed.

## II. RELATED WORK

The work on image recognition of multiple labeled dishes [1] uses four detectors to detect the candidate regions of the image and then the candidate regions are fused. Aguilar et al. [2] fine-tuned the object detection algorithm YOLOv2 [9] to perform multiple food detection and recognition. Zhang et al. [4] proposed a multi-task learning method that can use CNNs to simultaneously recognize the type of food, ingredients and cooking method from dish images. Ming et al. [12] used ResNet [14] network to directly extract visual features for dish image recognition. McAllister et al. [16] directly extracted visual features using pre-trained ResNet-152 and GoogleNet [17], and then used the fused deep features to train a classifier for dish image recognition. Mejía et al. [18] simplified the Mask R-CNN network [19] so that it focuses only on the dish image recognition task. Qiu et al. [31] proposed PAR-Net to mine discriminative food regions to improve the performance of classification. E. Aguilar et al. [30] proposed a new approach for food analysis based on CNN which integrates food localization, recognition and segmentation in the same framework. Min et al. [3] proposed the Deep Progressive Region Enhancement Network, which utilizes a self-attention mechanism to incorporate rich contextual information at multiple scales into local features to enhance the feature representation and thus improve the recognition performance.

Most works only adjust a single model. In contrast, our DEIHDL model utilizes differential evolution techniques to optimize weight allocation and integrates three different types of base models for detection.

## III. METHODOLOGY

Our DEIHDL model (as shown in Figure 1) is divided into two parts: the first part introduces the development of three basic models, and the second part demonstrates the integration of these basic models.

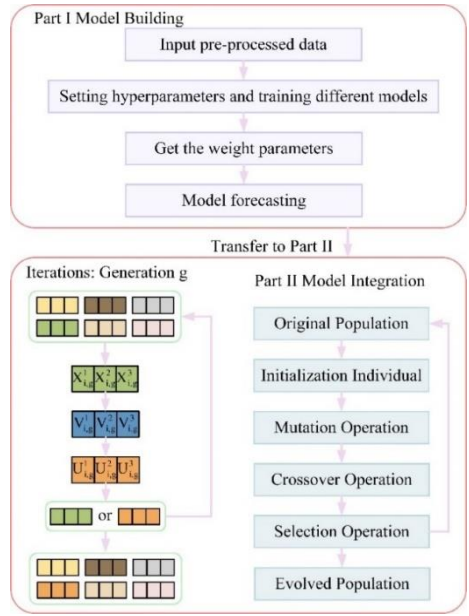

Fig. 1. The overall architecture of the DEIHDL.

### A. Model Building

#### 1) YOLOv5.

YOLOv5 [10] is the fifth version of the You Only Look Once (YOLO) series of algorithms, which is representative of single-stage object detection algorithms. It achieves high accuracy, efficiency and ease of use at that time.

It should be declared that S is the number of grid cells, B is the number of anchors on each grid cell, and IoU is the overlapping rate of the two boxes. The loss function as follow:

$$\mathcal{L} = \lambda_{box} \sum_{i=0}^{S^2} \sum_{j=0}^{B} \mathbb{I}_{i,j}^{obj} \left( 1 - IoU + \frac{\rho^2}{c^2} + \alpha v \right) + \lambda_{obj} BCE_{obj}^{sig}(conf, conf_{iou}) + \lambda_{cls} BCE_{obj}^{sig}(\hat{cls}, cls) \quad (1)$$

where $\lambda_*$ is the weight of the corresponding term, $\mathbb{I}_{i,j}^{obj}$ indicates whether the j-th anchor box of the i-th cell is a positive sample, $\rho$ is the diagonal length of the smallest enclosing box covering the predicted and ground truth, $v$ is a measure of the difference in aspect ratio between the predicted box and the ground truth, $\alpha$ is the balancing factor, $c$ is the length of the diagonal, $BCE(\cdot,\cdot)$ is the binary cross entropy function, $cls$ is the category score, and $conf$ is the confidence score.

*2) YOLOv8.*

YOLOv8 [11] can be used for various object detection and tracking, instance segmentation, image classification, and pose estimation tasks. The loss function is specified as follows:

$$\mathcal{L} = \frac{\lambda_{box}}{N_{pos}} \sum_{x,y} \mathbb{I}_{c_{x,y}^*} \left[ 1 - q_{x,y} + \frac{\|b_{x,y} - \hat{b}_{x,y}\|_2^2}{\rho^2} \alpha_{x,y} v_{x,y} \right] + \frac{\lambda_{cls}}{N_{pos}} \sum_{x,y} \sum_{c \in classes} \left( y_c \log(\hat{y}_c) + (1 - y_c) \log(1 - \hat{y}_c) \right)$$
$$+ \frac{\lambda_{dfl}}{N_{pos}} \sum_{x,y} \mathbb{I}_{c_{x,y}^*} \left[ -(q_{(x,y)+1} - q_{x,y}) \log(\hat{q}_{x,y}) + (q_{x,y} - q_{(x,y)-1}) \log(\hat{q}_{(x,y)+1}) \right] \quad (2)$$

where $q_{x,y}$ is the IoU between the prediction box and the ground truth. $N_{pos}$ is the total number of cells containing an object. $b_{x,y}$ denotes the center point of the ground truth. $\mathbb{I}_{c_{x,y}^*}$ is the indicator function that indicates the inclusion of the object. $v_{x,y}$ is calculated as follows:

$$v_{x,y} = \frac{4}{\pi^2} \left( \arctan\left(\frac{w_{x,y}}{h_{x,y}}\right) - \arctan\left(\frac{\hat{w}_{x,y}}{\hat{h}_{x,y}}\right) \right)^2 \quad (3)$$

where $w_{x,y}$ and $h_{x,y}$ are the width and height of the box respectively.

*3) DETR.*

The DETR model [13] is Transformer's groundbreaking work in the field of object detection. It simplifies the traditional detection process. DETR infers a fixed-size set of $N$ predictions, in a single pass through the decoder. To find a bipartite match between the ground truth set and the prediction set, DETR searches for the least costly permutation of $N$ elements $\sigma$:

$$\sigma = \arg\min \sum_i^N \mathcal{L}_{match}(y_i, \hat{y}_{\sigma(i)}) \quad (4)$$

where $\mathcal{L}_{match}(y_i, \hat{y}_{\sigma(i)})$ is a pair-wise matching cost between the ground truth $y_i$ and the prediction with index $\sigma(i)$.

$$\mathcal{L}_{match}(y_i, \hat{y}_{\sigma(i)}) = -\mathbb{I}_{\{c_i \neq \varnothing\}} p_{\sigma(i)}(c_i) + \mathbb{I}_{\{c_i \neq \varnothing\}} \mathcal{L}_{box}(b_i, \hat{b}_{\sigma(i)}) \quad (5)$$

where $p_{\sigma(i)}(c_i)$ is the probability of class $c_i$, $\hat{b}_{\sigma(i)}$ is he prediction box for predictions at index $\sigma(i)$. DETR defines $\mathcal{L}_{box}(b_i, \hat{b}_{\sigma(i)})$ using a linear combination of the generalized IoU loss [26] $\mathcal{L}_{iou}(\cdot,\cdot)$ and the $\mathcal{L}_1$ loss:

$$\mathcal{L}_{box}(b_i, \hat{b}_{\sigma(i)}) = \lambda_{iou} \mathcal{L}_{iou}(b_i, \hat{b}_{\sigma(i)}) + \lambda_{\mathcal{L}1} \|b_i - \hat{b}_{\sigma(i)}\|_1 \quad (6)$$

where $\lambda_{iou}, \lambda_{\mathcal{L}1} \in R$ are hyperparameters.

DETR uses the Hungarian loss function:

$$\mathcal{L}_{Hungarian}(y, \hat{y}) = \sum_{i=1}^N \left[ -\log p_{\sigma(i)}(c_i) + \mathbb{I}_{\{c_i \neq \varnothing\}} \mathcal{L}_{box}(b_i, \hat{b}_{\sigma(i)}) \right] \quad (7)$$

where $\sigma$ is the optimal allocation calculated from (4).

*B. Model Integration*

Differential evolution (DE) is used in model integration to optimize the self-tuning weights, including four operations such as initialization, mutation, crossover and selection. DE is commonly used for automated optimization tasks. D. Wu et al. [8] introduced hierarchical DE in model training to automatically determine two core hyperparameters. D. Wu et al. [15] developed a classification and localization optimization algorithm that combines DE.

*1) Initialization.*

Initializing a population of NP individuals. An individual $X_{i,g}$ is defined as follows:

$$X_{i,g} = [X_{i,g}^1, X_{i,g}^2, X_{i,g}^3] \quad (8)$$

where $i \in \{1, 2, ..., NP\}$ stands for the i-th individual; $g \in \{1, 2, ..., G\}$ stands for the g-th generation; $X_{i,g}^1$, $X_{i,g}^2$, and $X_{i,g}^3$ represent the integration weights of different models, respectively.

*2) Mutation.*

It uses DE/Rand/1 to generate a mutant individual $V_{i,g}$ for each target individual $X_{i,g}$ in the current generation:

$$V_{i,g} = X_{r1,g} + F_i \cdot (X_{r2,g} - X_{r3,g}) \quad (9)$$

where $X_{r,g}$ is different individual randomly selected from NP individuals; $F_i$ is a scaling factor. The Scaling Factor Local Search DE (SFLSDE) method [20] can effectively achieve $F_i$ adaptation:

$$F_i = \begin{cases} SFGSS & \text{if } rand_3 < \tau_2 \\ SFHC & \text{if } \tau_2 \leq rand_3 < \tau_3 \\ F_l + F_u \cdot rand_1 & \text{if } rand_2 < \tau_1 \text{ and } rand_3 > \tau_3 \\ F_i & \text{otherwise} \end{cases} \quad (10)$$

where $rand$ is uniform pseudorandom number between 0 and 1; $\tau$ is constant threshold value.

*3) Crossover.*

An arithmetic crossover strategy is used to perform crossover operations on the target individual and their corresponding mutant individual to generate new trial individual:

$$U_{i,g} = X_{i,g} + K \cdot (V_{i,g} - X_{i,g}) \quad (11)$$

where $K$ is a random number from 0 to 1.

*4) Selection.*

The selection operation determines which individual will be retained into the next generation. The following is how the initial individuals of the next generation are selected:

$$X_{i,g} = \begin{cases} U_{i,g} & \text{if } Performance(U_{i,g}) > Performance(X_{i,g}) \\ X_{i,g} & \text{otherwise} \end{cases} \quad (12)$$

$$Performance(X_{i,g}) = Evaluate(WBF(X_{i,g})) \quad (13)$$

where $Performance(X)$ represents the performance of individual weights merged on the validation set; $Evaluate(\cdot)$ is a function to measure the performance of deep learning (DL)-based object detection models based on specific metrics; $WBF(\cdot)$ is a weighted fusion bounding box strategy proposed by Solovyev R et al. [23] that combines object detection models predictions.

## C. Algorithm design

We also design the algorithm DEIHDL, and the pseudo-code is shown in Table I. The pseudo-code of the WBF strategy is shown in Table II.

The computational complexity of WBF is:

$$T_{WBF} = O(N_B) + N_B \times O(1) + N_F \times O(1) = O(N_B) \quad (14)$$

Combining Table I and Table II, the computational complexity of the DEIHDL algorithm is:

$$T_{DEIHDL} = NP \times O(1) + G \times NP \times O(N_B) = O(G \times NP \times N_B) \quad (15)$$

TABLE I. ALGORITHM DEIHDL.

```
Algorithm 1: DEIHDL
  Input: G, NP
  Output: best individual X_{best,g}
1 begin
2   g = 0                                                        O(1)
3   for i ← 1 to NP do                                           ×NP
4     | randomly generate and initialize population individual X_{i,g}  O(1)
5   end for
6   while g < G do                                               ×G
7     for i ← 1 to NP do                                         ×NP
8       randomly select X_{r1,g}, X_{r2,g}, and X_{r3,g} in population  O(1)
9       compute V_{i,g} by Eq.(9)                                O(1)
10      generate U_{i,g} from X_{i,g} and V_{i,g} by Eq.(11)     O(1)
11      Performance_{X_{i,g}} = Evaluate(WBF(X_{i,g}))           O(N_B)
12      Performance_{U_{i,g}} = Evaluate(WBF(U_{i,g}))           O(N_B)
13      if Performance_{U_{i,g}} > Performance_{X_{i,g}} then
14        | X_{i,g} = U_{i,g}                                    O(1)
15      end if
16    end for
17    g = g + 1                                                  O(1)
18  end while
19  return X_{best,g}
20 end
```

TABLE II. FUNCTION WBF.

```
Algorithm 2: WBF(X_{i,g})
  Output: fusion box list F
1 begin
2   summarize the prediction boxes from all models into list Box    O(1)
3   sort boxes in Box by confidence score                           O(N_B)
4   set the W of each box in Box by Eq.(18)                         O(1)
5   declare empty lists L and F                                     O(1)
6   for b_i in Box do                                               ×N_B
7     if no matching box in F then
8       F ← F ∪ b_i                                                 O(1)
9       L[end] ← L[end] ∪ b_i                                       O(1)
10    end if
11    if find matching box in F at pos then
12      L[pos] ← L[pos] ∪ b_i                                       O(1)
13      compute F[pos] by Eq.(14)(15)(16)                           O(1)
14    end if
15  end for
16  for f_i in F do                                                 ×N_F
17    compute confidence score for each category of f_i by Eq.(17)  O(1)
18    select the category with the max confidence score             O(1)
19  end for
20  update confidence score by Eq.(19)                              O(1)
21  return F
22 end
```

## IV. EXPERIMENTS AND RESULTS

In the subsequent experiments, we aim at answering the following research questions (RQs):

RQ. 1. Is the DEIHDL model superior to individual model?

RQ. 2. What hyperparameters affect the performance of DEIHDL model?

### A. General Settings

**Datasets.** The experiment is based on the real dataset Dish Ingredients, which was collected from the Internet and consists of 11 ingredient categories. Table III summarizes its statistics.

TABLE III. STATISTICS OF THE STUDIED DATASETS.

| Dataset Name | Statistics | | |
| --- | --- | --- | --- |
| | Classes | Images count | Instances count |
| Dish Ingredients | 11 | 2200 | 4609 |

**Evaluation Metrics.** We use the following evaluation metrics:

$$Precision = \frac{TP}{TP + FP}$$

$$Recall = \frac{TP}{TP + FN}$$

$$AP = \int_0^1 P(R) d(R)$$

$$mAP = \frac{1}{classes} \sum_{i=1}^{classes} AP_i$$

### B. Performance Comparison (RQ. 1)

We compared DEIHDL with six models. Their details are summarized in Table IV.

TABLE IV. SUMMARY OF COMPARISON MODELS.

| Model | Description |
| --- | --- |
| YOLOv5 [10] | A classical single-stage object detection model. It improves accuracy while maintaining real-time detection speed. |
| YOLOv8 [11] | It is a single-stage object detection model. It uses an Anchor-Free detection head to speed up inference. |
| DETR [13] | An object detection model based on the Transformer architecture. It simplifies the traditional detection process. |
| Gold-YOLO [5] | An efficient real-time object detection model. It enhances multi-scale feature fusion through the Gather-and-Distribute (GD) mechanism. |
| DAMOYOLO [6] | It is an efficient object detection model. It achieves industry-leading detection accuracy while maintaining fast inference speed. |
| YOLOX [7] | A classical single-stage object detection model. It introduces a decoupled head and a leading label assignment strategy. |

The experimental data (Table V) show that DEIHDL outperforms a single model. In short, the integrated approach enhances the effectiveness of DEIHDL and shows its strong potential as a prediction tool that can bring more accurate results for multiple applications.

TABLE V. THE COMPARISON RESULTS OF THE ACCURACY.

| Model | mAP50 (%) | mAP50-95 (%) |
|---|---|---|
| YOLOv5 | 82.96 | 62.14 |
| YOLOv8 | 87.55 | 71.01 |
| DETR | 86.00 | 67.20 |
| Gold-YOLO | 75.45 | 59.48 |
| DAMOYOLO | 86.08 | 70.91 |
| YOLOX | 88.41 | 68.27 |
| **DEIHDL** | **90.92** | **72.25** |

*C. Hyperparameter analysis (RQ. 2)*

In this experiment, when one hyperparameter was tested, the other parameters were set to default values.

*1) Number of individuals in the population.*

To explore the effect of NP on the effectiveness of DEIHDL, we implemented experiments with multiple NP configurations. As demonstrated in Fig. 2, we found that DEIHDL performs best when the NP is between 5 and 15.

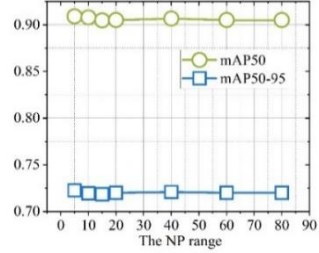

Fig. 2. The influence of NP on the dataset.

*2) Number of evolutionary generations.*

To explore the effect of generation size G on the effectiveness of DEIHDL, we implemented experiments with multiple G configurations. As demonstrated in Fig. 3, we found that DEIHDL performs best when the G is 40.

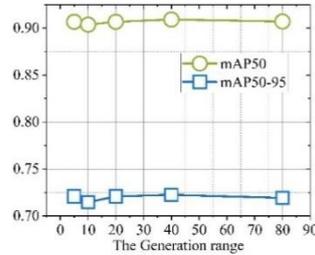

Fig. 3. The influence of G on the dataset.

*3) Different weights.*

The weighting curves in Fig. 4 illustrate the distribution of weights for each model, showing that the models tend to converge as the number of evolutionary generations increases.

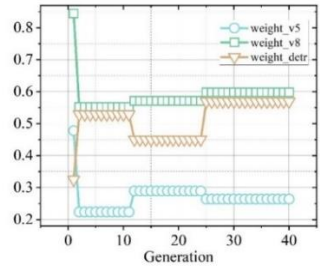

Fig. 4. The influence of weights on the dataset.

## V. Conclusion

In this study, the DEIHDL model is proposed to utilize the integration method for accurate object detection in a complex pre-made dishes scene, and its effectiveness is verified on a dataset. The experimental results show that the DEIHDL model can be adaptively optimized to achieve the best integration effect and improve the detection accuracy. However, the model is also limited by data integrity and hyperparameter tuning. There are some studies that can be used to ameliorate the above limitations: multi-metric latent feature analysis [21], representation and learning of incomplete data [22,51,72-75], outlier-resilient autoencoder for representing incomplete data [24], state-migration particle swarm optimizer that can adaptively latent factor analysis of incomplete data [25], pseudo gradient-adjusted particle swarm optimization algorithm [27,69-71], and other work on parameter optimization work [32-42,60-68].

In the future, we intend to utilize intelligent optimization algorithms to automatically tune hyperparameters and reduce the impact of data incompleteness in DEIHDL. Also, we will employ online semi-supervised learning [28] to mitigate the problem of under-labeled images.